\documentclass[letterpaper, 10 pt, conference]{ieeeconf}  %

\IEEEoverridecommandlockouts                              %

\usepackage[tracking=true]{microtype}
\usepackage{amsmath} %
\usepackage{amssymb}  %
\usepackage{graphicx}
\usepackage{epsfig} %
\usepackage{mathptmx} %
\usepackage{times} %
\usepackage{amsmath} %
\usepackage{amssymb}  %
\usepackage{xspace}
\usepackage[frozencache,cachedir=minted]{minted}
\setminted{fontsize=\footnotesize}
\usepackage{url}
\usepackage[dvipsnames]{xcolor}
\usepackage[font={footnotesize}]{caption}
\usepackage{booktabs} %
\usepackage{tikz}
\usetikzlibrary{positioning,shapes.geometric,arrows,fit,shapes.symbols,svg.path,tikzmark,calc}
\usepackage{booktabs}
\usepackage{textcomp}
\usepackage{listings}
\usepackage{subcaption}
\pgfdeclarelayer{background}
\pgfdeclarelayer{foreground}
\pgfsetlayers{background,main,foreground}
\usepackage{siunitx}
\usepackage{adjustbox}
\usepackage{array}
\usepackage{multirow}

\let\labelindent\relax %

\usepackage[inline]{enumitem} %

\newcommand{\algname}{FogROS2-SGC\xspace}
\newcommand{\proxy}{SGC proxy\xspace}
\newcommand{\router}{SGC router\xspace}

\newcommand{\remark}[3]{}
\newcommand{\jeff}[1]{\remark{0.0}{JI}{#1}}
\newcommand{\eric}[1]{\remark{0.85}{EC}{#1}}

\newcommand{\ken}[1]{\remark{0.05}{KG}{#1}}

\newcommand{\numberedparagraph}[1]{\textit{#1:} }

\title{\LARGE \bf
FogROS2-SGC: A ROS2 Cloud Robotics Platform \\ for Secure Global Connectivity 
}

\author{Kaiyuan Chen$^{1}$, Ryan Hoque$^{1}$,   Karthik Dharmarajan$^{1}$,    Edith LLontop$^{1}$,  Simeon Adebola$^{1}$,  \\
Jeffrey Ichnowski$^{2}$,  
  John Kubiatowicz$^{1}$, and Ken Goldberg$^{1,3}$%
\thanks{$^{1}$Department of Electrical Engineering and Computer Science}%
\thanks{$^{2}$Carnegie Mellon University, Pittsburgh, PA, USA}%
\thanks{$^{3}$Department of Industrial Engineering and Operations Research}%
\thanks{$^{1,3}$University of California, Berkeley, CA, USA }%
\thanks{{\tt\footnotesize kych@berkeley.edu}}%
}

\begin{document}

\maketitle
\thispagestyle{empty}
\pagestyle{empty}

\begin{abstract}
The Robot Operating System (ROS2) is the most widely used software platform for building robotics applications. FogROS2 extends ROS2 to allow robots to access cloud computing on demand. However, ROS2 and FogROS2 assume that all robots are locally connected and that each robot has full access and control of the other robots. With applications like distributed multi-robot systems, remote robot control, and mobile robots, robotics increasingly involves the global Internet and complex trust management. Existing approaches for connecting disjoint ROS2 networks lack key features such as security, compatibility, efficiency, and ease of use. We introduce \algname, 
an extension of FogROS2 that can effectively connect robot systems across different physical locations, networks, and Data Distribution Services (DDS). With globally unique and location-independent identifiers, \algname securely and efficiently routes data between robotics components around the globe. \algname is agnostic to the ROS2 distribution and configuration, is compatible with non-ROS2 software, and seamlessly extends existing ROS2 applications without any code modification. Experiments suggest \algname is 19$\times$ faster than rosbridge (a ROS2 package with comparable features, but lacking security).
We also apply \algname to 4 robots and compute nodes that are 3600 km apart. 
Videos and code are available on the project website \url{https://sites.google.com/view/fogros2-sgc}.

\end{abstract}

\section{Introduction}\label{sec:intro}

As robots are increasingly deployed worldwide, they require mechanisms to efficiently, reliably, and securely communicate with other robots, sensors, computers, and the cloud. The applications are broad, from mobile robots with changing IP addresses due to traveling through different networks, to a fleet of globally distributed robots learning collaboratively. 
Cloud and fog robotics~\cite{kehoe2015survey} empower robots and automation systems to harness off-board resources in cloud-based computers. %
In prior work, we introduced FogROS2~\cite{ichnowski2022fogros}, now an official part of the ROS2 ecosystem~\cite{fogros2_ros2_repo}, to enable robots to execute modern compute and memory-intensive algorithms using on-demand hardware resources on the edge and cloud. 
However, robots connecting to the cloud, a nearby computer on a different network, or a robot halfway around the world introduce additional challenges: 
(1) Making robots accessible to other robots on the public internet may leave them vulnerable to unauthorized connections and data breaches.
(2) The heterogeneity of interconnected devices, communication protocols, and configurations
causes incompatibilities that hinder integration and operation.
(3) The changing network topology of mobile robots and Unmanned Aerial Vehicles (UAVs) challenges their ability to stay connected. 
To illustrate some of these challenges (Fig.~\ref{fig:examples}), consider:

\begin{figure}
    \centering
    \includegraphics[width=0.95\linewidth]{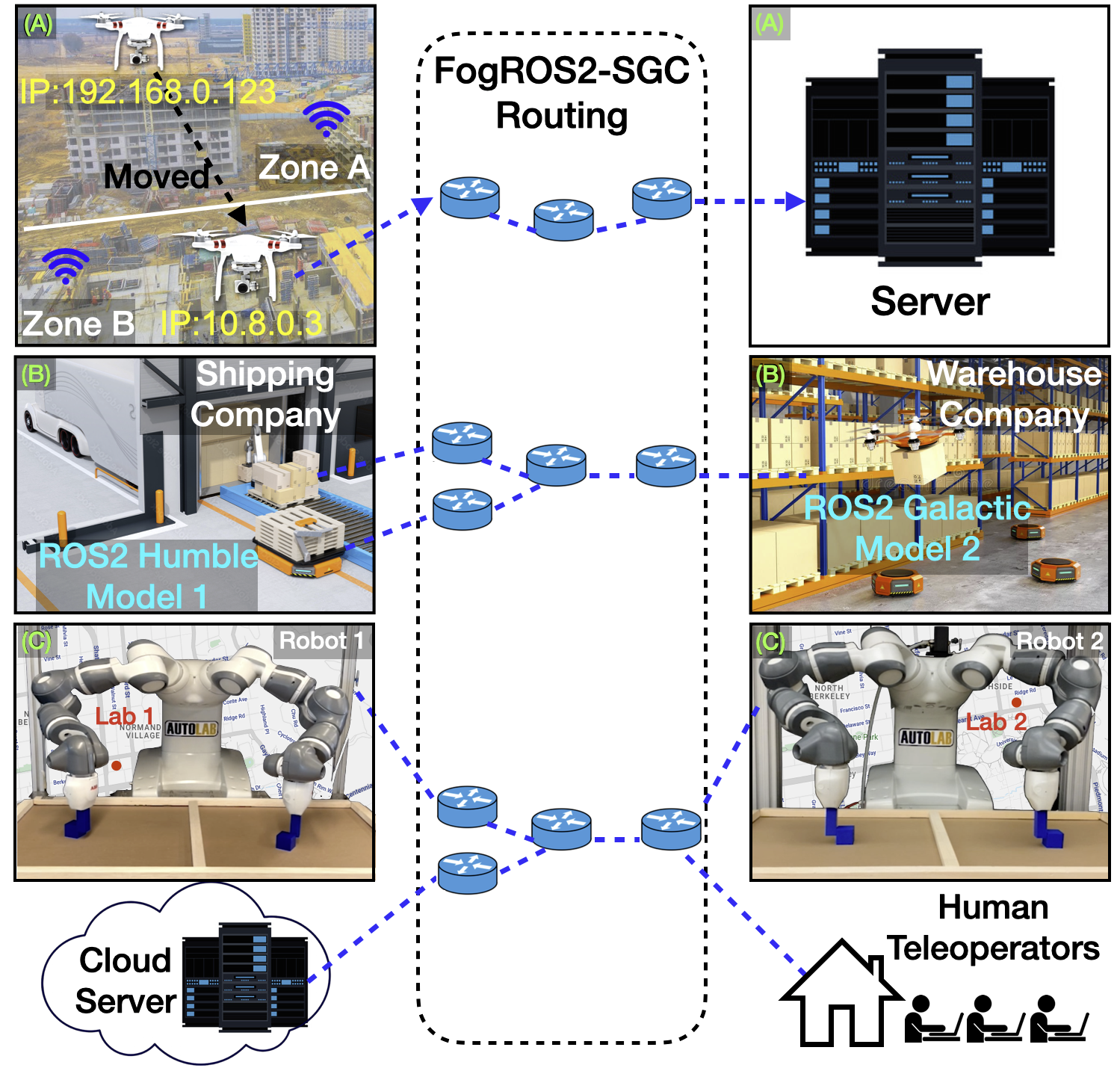}
    \caption{\algname{} enables \underline Secure \underline Global \underline Connectivity for robots, allowing robots to communicate with other robots, computers, and the cloud through a familiar ROS2 interface.  With \algname{}, (A) drones navigating large construction sites can seamlessly communicate, even when their IP addresses are constantly changing due to switching Wi-Fi and cellular networks; (B) shipping and stocking robots from different corporations can securely share only the required topics necessary to facilitate the transfer of goods at a warehouse; and (C) globally distributed robots can participate in fleet learning.  In experiments, we demonstrate \algname{} on Fleet-DAgger~\cite{hoque2022fleet}, a fleet learning algorithm, with 4 robot arms operating simultaneously in different locations.%
    }
    \label{fig:examples}
    \vspace{-15pt}
\end{figure}

\numberedparagraph{(A) Security and inspection drones} %
Drones navigate a construction site and stream data to a central station that updates a dynamically-changing SLAM map~\cite{openvslam2019}. As drones fly through different cellular and Wi-Fi networks, their IP addresses change, but they should remain securely connected. %

\numberedparagraph{(B) Coordinating heterogeneous mobile robots in a warehouse}
Robots belonging to different companies (e.g., shipping vs. warehouse) and of different makes and models hand off items between container and warehouse. %
Each robot has unique software packages and versions (e.g., operating systems and network protocols) and 
must communicate, but only a few selected topics are necessary for the handoff. %

\numberedparagraph{(C) Distributed fleet learning}
Robot arms at different locations %
pool their data and collectively update a shared control policy. Robots unable to make progress can fall back on remote human teleoperators, using algorithms such as Fleet-DAgger~\cite{hoque2022fleet}.

To address these challenges, we present \algname (Secure Global Connectivity), an extension of FogROS2 that securely and reliably connects robots across different software components, network protocols, and physical locations. 
\algname enables disjoint ROS2 networks to connect to ROS2 topic interfaces named with \textit{globally-unique} and \textit{location-independent} identifiers. 
The robots using \algname can  roam freely while staying connected because the identifiers are constant.  The identifiers are 256-bit strings that are secure by construction---a brute-force attack would have to find a match among $10^{77}$ possibilities (a value close to the number of protons in the observable universe\footnote{\vspace{-6pt}The Eddington Number~\cite{eddington1923mathematical} ($N_{\mathrm{Edd}}$) is currently estimated to be $10^{80}$.}).
\algname adopts a \textit{security-first} routing design, where only authenticated parties can connect to the robot and establish secure communication. In contrast to prior work such as SROS2~\cite{mayoral2022sros2} and FogROS2~\cite{ichnowski2022fogros}, \algname does not require merging distributed ROS2 networks, allowing robots to keep their ROS2 networks private and expose public topics only if explicitly configured. Providing fine-grain isolation and access control reduces the attack surface and enhances scalability.

\algname seamlessly integrates with ROS2 applications without code modifications via an \proxy. 
Its implementation and security policy configuration are agnostic to ROS2 distributions and their network transport middleware vendors. %
\algname is also compatible with non-ROS2 programs that interact with ROS2 components and can provide secure global connectivity to non-cloud servers and computers. Furthermore, since memory copy and synchronization operations are expensive for memory-constrained robots, the implementation of \algname processes can route data without performing unnecessary copies %
(also known as ``zero copy'').

Experiments suggest that \algname reduces the network latency of a cloud-based grasp planning application by 9.42$\times$ compared to unsecured rosduct~\cite{rosduct}-rosbridge~\cite{crick2012rosbridge}.  
We also deploy \algname to simultaneously control a fleet of 4 robot arms in different physical locations with compute off-loaded to a server 3600 km away.

This paper makes the following contributions: 
\begin{enumerate*}[label=(\arabic*)]
\item \algname, an extension of FogROS2 that connects disjoint ROS2 networks by assigning public ROS2 topics with globally-unique and location-independent identifiers.
\item Method for secure and efficient routing with \algname.  
\item A Rust implementation of \algname that uses zero-copy message processing and asynchronous network operations for robots with memory and compute constraints.
\item Evaluation of \algname on cloud robotics applications (vSLAM, grasp planning, motion planning, simultaneous fleet control) demonstrating up to 9.42$\times$ latency reduction and enhanced usability.  \jeff{X scenarios demonstrating...} \eric{what number should we include? should we merge with the previous paragraph? } 
\end{enumerate*}

\begin{figure}
    \centering
    \includegraphics[width=\linewidth]{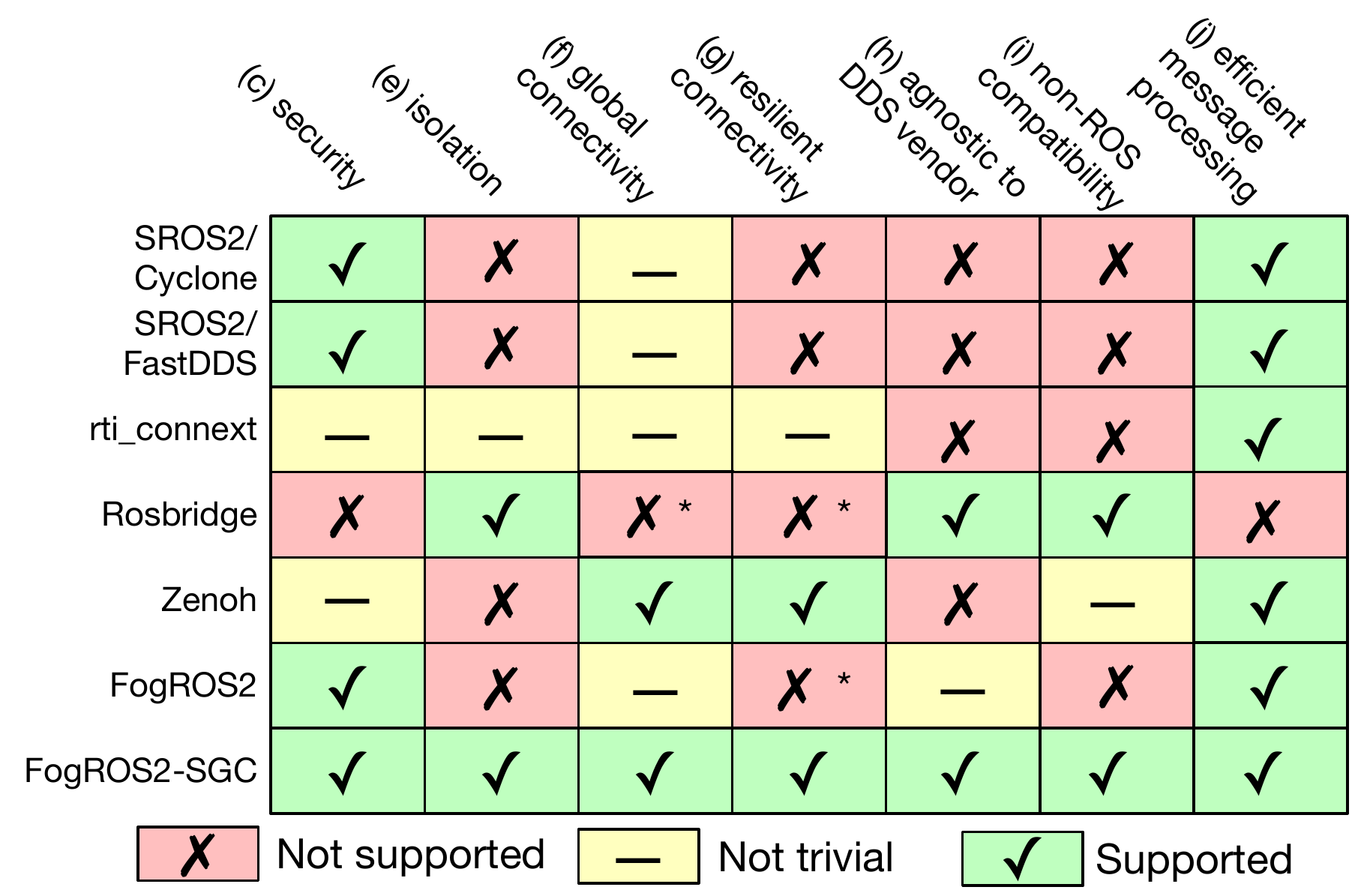}
    \caption{\textbf{Comparison of \algname with other distributed ROS2 systems.} In this table, we compare the feature support of different distributed ROS2 systems with the features in Section~\ref{sec:intro:principle}. Some features can be supported but require non-trivial effort beyond changing the configuration. For example, both the routing service in rti\_connext and discovery server in FastDDS/SROS2 support global connectivity but require manually modifying routing rules or setting up a point-to-point VPN when a new node joins ~\cite{fastdds_problem}. Rosbridge and FogROS2 support only unidirectional global and resilient connectivity (marked with *), meaning that one side of the communication must have a fixed IP. In contrast, the identifier-based routing of \algname allows either side to have a dynamic IP address. }
    \label{fig:feature_table}
    \vspace{-15pt}
\end{figure}

\section{Related Work}

James Kaufner introduced the term 'Cloud Robotics' in  2010 \cite{kehoe2015survey}. Cloud and fog computing have been applied to robotic tasks such as grasp planning (Tian et al.~\cite{tian2017cloud}, Kehoe et al.~\cite{kehoe2013cloud}, and Li et al.~\cite{li2018dex}), parallelized Monte-Carlo grasp perturbation sampling (Kehoe et al. \cite{kehoe2012estimating,kehoe2012toward,kehoe2014cloud}), and motion planning (Lam et al.~\cite{lam2014path}). 
Chen et al.~\cite{chen2021fogros} and Ichnowski et al.~\cite{ichnowski2022fogros}. propose frameworks for offloading computation to resources on the edge or cloud, while Ichnowski et al. ~\cite{ichnowski2020fog} and Anand et al.~\cite{anand2021serverless} present systems that leverage serverless computing~\cite{mcgrath2017serverless}. 
Modern computing paradigms have enabled new applications such as multi-robot interactive fleet learning (Swamy et al.~\cite{swamy2020scaled}, Hoque et al.~\cite{hoque2022fleet}) and remote sharing of robot systems (Tanwani et al. ~\cite{tanwani2020rilaas}, Bauer et al. \cite{robotcluster}). 

Remote interactions between robots and the cloud raise security, compatibility, and connectivity challenges for robots. 
Virtual Private Networks (VPNs) are the most common approach for establishing secure communication between robots and the cloud for both ROS and ROS2 (e.g., Lim et al. \cite{lim2019cloud}). Establishing a VPN link between a robot and the cloud is a complex process~\cite{hajjaj2017establishing}.
FogROS~\cite{chen2021fogros} and FogROS2~\cite{ichnowski2022fogros} automate the certificate generation and VPN setup.
SROS2~\cite{mayoral2022sros2} is an alternative approach to securing ROS2 communication that enforces access control of ROS2 topics. However, it requires DDS-dependent discovery mechanisms to ensure connectivity. 
Discovery mechanisms for DDS (such as the discovery server for FastDDS~\cite{fastdds} and the RTI routing service for RTI Connext~\cite{connext}) are vendor-specific and not compatible with other DDS implementations. 
Zenoh for ROS2~\cite{zenohros2} is integrated with CycloneDDS to enhance peer-to-peer connectivity, but it is not compatible with other DDS implementations. 
ROS Remote~\cite{pereira2019rosremote} by Pereira et al. and  MSA~\cite{xu2020cloud} by Xu et al. propose alternative protocols to unify cloud-robot communication. However, alternative protocols require modifications to ROS applications and are not compatible with ROS2. Finally, rosbridge~\cite{crick2012rosbridge} proposed by Crick et al. is widely adopted by both ROS1 and ROS2 to allow non-ROS software to interact with ROS2 nodes. It can also be used to bridge two non-compatible and remote ROS applications when used in conjunction with rosduct~\cite{rosduct}. However, rosduct and rosbridge have significantly high message latency when the message size is large (e.g., images). A summary of how \algname differs from related work can be found in Fig. \ref{fig:feature_table}.

\section{Ten \algname Features}
\label{sec:intro:principle}
\algname addresses the Secure Global Connectivity (SGC) problem of securely and reliably connecting globally distributed robots, sensors, computers, and the cloud by extending FogROS2. We enumerate 10 new features to differentiate from related libraries and alternative approaches. %
\paragraph{Globally identifiable addresses} %
\label{sec:intro:global}
\algname enables a scalable number of ROS2 networks to publish a subset of ROS2 topics to other disjoint ROS2 networks around the globe. 
In scenario (C) from Fig.~\ref{fig:examples}, the robot arms are located at different geographic locations with different local ROS2 networks. 
\algname allows remote human teleoperators to operate the robot arms as if the arms are connected to local networks. 
\algname also allows disjoint robots to publish to the same local ROS2 topics with globally unique and identifiable addresses. 
\ken{remove FogROS2 also allows
...} \eric{we want to highlight global addressibility here}

\paragraph{Transparency to ROS2 applications}
ROS2 modularizes a robotics application into \textit{nodes}, and connects the nodes into a graph. Nodes communicate with each other through a publish-subscribe (pub/sub) system, where publisher nodes send messages to \textit{topics}, and nodes subscribed to these topics receive these messages. 
\algname adheres to the abstractions and interfaces of ROS2.  ROS2 applications interact with remote nodes as if they are nodes on the same robot or subnetwork. 

\paragraph{Communication security}
\algname guarantees that no unauthorized attacker can eavesdrop or tamper with ROS2 messages. Authorization is identified by user-configured cryptographic keys. In all three scenarios from Section~\ref{sec:intro}, the communication goes through wide-area network with untrusted infrastructure. \algname prevents attackers from accessing any content in ROS2 messages and differentiates authentic robots from spoofing attackers. 

\paragraph{Global anonymity}
Authorized participants can deterministically derive global identifiable addresses with ROS2 topic information and cryptographic secrets.
Attackers cannot reverse any information used to recover addresses or topics. 
\algname guards against attackers knowing part of the ROS2 topic information deducing the global address. For example, the attackers who know the topic name and type information cannot guess the address, because they miss the author information and security credentials of ROS2 node. \jeff{I'm not sure this is a sufficient requirement.  I think pseudo-random number generators have the same property, and are \emph{not} secure.}

\paragraph{ROS2 network isolation and topic-level access control}
\algname connects robots without merging distributed ROS2 networks. 
Every robot can have an arbitrary number of private ROS2 topics and only public interfaces are shared with other authorized ROS2 networks. Other ROS2 nodes interact with these public interfaces just as they interact with a local ROS2 topic. \jeff{Have we defined `public topics?'  I'm not sure what it means.} \eric{how about now?}
This protects the privacy of the robot and prevents unintended messages from being shared with other disjoint networks of the system. For example, in scenario (B), a delivering robot from one company and receiving robot from another may have some proprietary topics that are kept private from each other. \algname isolates the topics private to the robots.

\paragraph{Global connectivity}
Some robots are connected to subnetworks that are not directly accessible from the outside. For example, robots in scenario (C) are in local area networks behind Network Address Translation (NAT). NAT allows multiple robots to share the same IP, but the translation is dynamic and ROS2 nodes outside cannot directly access the robots. \algname can connect ROS2 nodes that are behind firewalls and NAT.

\paragraph{Seamless and resilient connectivity to network dynamism} 
\algname adapts to the dynamic network behaviors of drones and mobile robots. %
\algname does not rely on static IP addresses to identify the robots because such addresses are usually bound to a physical location. Adding or reconnecting to robots should not restart the ROS2 node entirely, as this causes service interruptions and failures. 
 
\paragraph{DDS-agnostic compatibility}
ROS2 adopts the Data Distribution Service (DDS) as its underlying network transport middleware to marshal, unmarshal, and exchange messages. 
ROS2 supports different DDS implementations, such as CycloneDDS~\cite{cyclonedds}, FastDDS~\cite{fastdds}, and RTI Connext~\cite{connext}. However, a warehouse in scenario (B) may have robots running different versions of ROS2 and DDS. \algname is DDS-agnostic by leveraging ROS2 abstractions and not using any DDS-specific interfaces. 

\paragraph{Compatibility with non-ROS2 software}
\algname allows non-ROS2 software to interact with ROS2 nodes.  This principle is inspired by rosbridge~\cite{crick2012rosbridge}.
Besides common transport protocols,  while ROS2 officially supports C++ and Python, \algname allows
programs to use gRPC~\cite{grpc}, the most widely used Remote Procedure Call (RPC) framework that can run in heterogeneous environments and major popular programming languages (such as Go, Java, Javascript, PHP, Rust), to control the robots. \jeff{``any'' here would seem to overclaim about gRPC's capabilities---is this true?  Can we restate to something like ``popular'' languages?} \eric{that's what gRPC's website says...; how about now?} 

\paragraph{Efficient message processing and routing}
ROS2 messages are buffered in memory to be processed by \algname. Because robots often have memory and compute resource constraints, \algname is memory-efficient by reducing unnecessary message copying and memory synchronization. Since \algname requires frequent exchange of packets to and from the network, network operations (such as \texttt{send} and \texttt{recv}) are not on the critical path of message processing. 

\section{\algname Design}

\begin{figure}
    \centering
    \includegraphics[width=\linewidth]{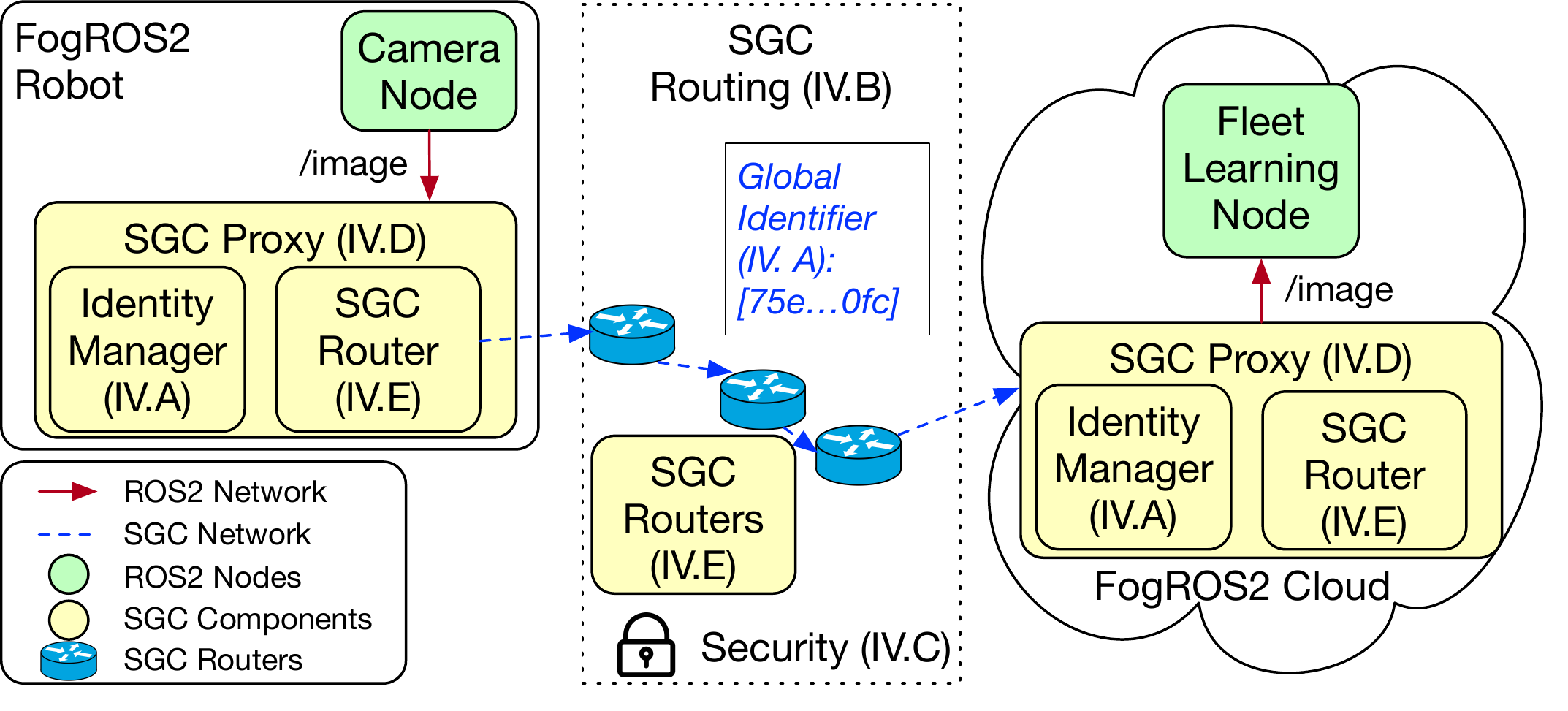}
    \caption{\textbf{System overview of \algname's architecture} showing a connection between a robot camera stream (on the ROS2 topic \texttt{/image}) and the cloud. The \algname{} assigns the ROS2 topic \texttt{/image} an anonymous, globally-unique and location-independent 256-bit identifier \texttt{[75e...0fc]} (truncated for brevity). The messages between identifiers are securely routed with \router.  }
    \label{fig:design:overview}
    \vspace{-15pt}
\end{figure}

See (Fig.~\ref{fig:design:overview}) for system architecture. \algname sends messages via a globally unique identifier (Sec.~\ref{sec:design:name}).  This identifier is unique to a robot and topic pair; thus, it can be used for sending and receiving messages regardless of robot location or network address (Sec.~\ref{sec:design:routing}).  The identifier is secure, and communication is encrypted, meaning only authorized robots and nodes can access messages from its referenced topic (Sec.~\ref{sec:design:secure}).  To implement routing based on the identifier, \algname consists of two main software components---(1) a router (Sec.~\ref{sec:design:routing} and \ref{sec:design:secure}), responsible for securely routing messages between other routers and nodes, and (2) a proxy (Sec~\ref{sec:design:proxy}), that converts between ROS2 messages and the secure routers.   As robots can be compute- and memory-constrained, \algname provides an efficient implementation (Sec.~\ref{sec:design:router}).

\subsection{Global Addressability}
\label{sec:design:name}
Maintaining a globally unique identifier enables the identification of a specific robotic component across subnetworks. 
\algname uses ROS2 topics as the minimal granularity for the global identifier because a topic is an interface to ROS2 nodes, and a ROS2 node can publish or subscribe to multiple ROS2 topics at the same time. For example, a ROS2 vSLAM node in openVSLAM~\cite{openvslam2019} has four ROS2 topics for camera information, video streaming, output localization, and mapping information. These
ROS2 topics expose standardized interfaces with fixed message types.
Users can limit the exposure of the ROS2 network by allowing only parts of the interface to be public. Partitioning public and private interfaces also enhances privacy and isolation, prevents unintended message exchanges, and reduces communication overhead. 

The identifier is designed to be unique, deterministic, and location-independent. 
To avoid name collisions, every identifier has 256 binary bits, leading to $2^{256}$ possible identifiers. 
Instead of letting users decide, all identifiers are cryptographically derived from the metadata of the ROS2 topics by an identifier manager in \proxy. The \proxy collects metadata such as the ROS2 node's name, author, maintainer, interface, and description from standard ROS2 interface and user configuration file. 
The metadata also has a unique string in case the user needs to deploy the same topic at different locations.
Every topic has an associated security certificate in X.509~\cite{myers1999internet} to verify the identity of those who want to publish or subscribe to the network.
All the metadata is serialized and converted into a 256-bit string using SHA-256~\cite{dtls_algorithm}, a widely used cryptographic hashing algorithm that maps arbitrary lengths of text to almost-unique 256-bit binary strings. 

\textbf{Security Analysis}: The hashed string is suitable for use as the globally unique identifier for the followingreasons: (1) \textbf{Deterministic:} The hash is deterministic so that every party holding the same metadata can derive the same hash value and thus the same global identifier. (2) \textbf{One-way:} SHA-256 is a one-way function, so the attacker cannot deduce or reverse the original metadata from the 256-bit identifier. (3) \textbf{Avalanche effect:} A small change to the original metadata leads to a new hash value that appears unrelated to the original hash value. (4) \textbf{Large namespace:} There are $2^{256}$ possible identifiers and it has been proved to be computationally intractable to find two messages with the same hash. Verification of these guarantees can be found in Appel~\cite{appel2015verification}.

\subsection{Location-Independent Routing}
\label{sec:design:routing}
Although having all identifiers in the same globally-flat namespace protects the privacy of the node's identity information and physical location, the identifiers do not carry any routing information. Flipping a bit in the identifier may lead to a drastic change in its physical location, or from existent to nonexistent. Therefore, securely routing messages between flat identifiers is a challenging problem. 
To solve this problem, \algname consolidates and extends the Global Data Plane (GDP)~\cite{mor2019global}, a peer-to-peer network that routes messages between location-independent identifiers. 
The routers are set up by the user and peer-wise connected into a routing graph; robots do not need to know other robots' addresses as long as there is a connected routing path. The routers can be any machine that has network and general compute capabilities, such as an edge computer or a cloud server. 
Every router stores the mapping between the identifiers and the corresponding routing information of the identifiers in Routing Information Base (RIB).

A joining robot or router broadcasts an \textit{advertisement} packet that announces the existence of the identifier and the routing information to the robot. The packet format is aligned with other \algname packets in Fig. \ref{fig:design:header}. Other routers store the routing information in RIB and broadcast the advertisement packet. Routing is achieved by looking up the destination routing information in the RIB and forwarding to that destination.

\begin{figure}
    \centering
    \includegraphics[width=\linewidth]{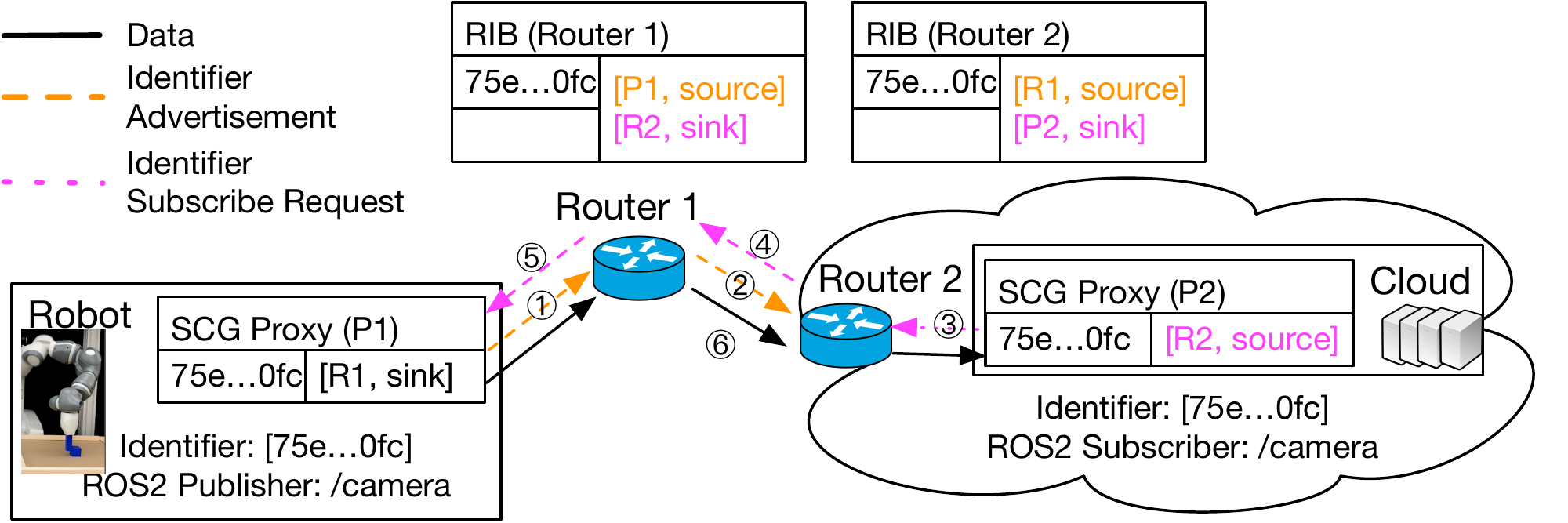}
    \caption{An illustration of how a routing connection is established between robot and cloud. The steps are further described in Section \ref{sec:design:routing}. (1,2) Advertisement generation and publish. (3,4,5) Subscribe request. (6) Data routing.}
    \label{fig:gdp_overview}
    \vspace{-10pt}
\end{figure}

Fig. \ref{fig:gdp_overview} illustrates a step-by-step example of a publishing and subscribing \texttt{/camera} topic with \algname. The figure assumes that all the connections between routers are established. This can be achieved through configuration or dynamic node discovery~\cite{mdns}.  The steps are:
\begin{enumerate*}[label=(\arabic*)]
    \item The robot \proxy P1 generates an advertisement message for the ROS2 topic \texttt{/camera} and sends it to Router 1. 
    \item After verifying the advertisement message, Router 1 records the advertisement in its RIB and forwards the topic information to Router 2. There can be multiple routers between Router 1 and Router 2. 
    \item Cloud \proxy P2 requests to subscribe to \texttt{/camera}, and the subscribe request is sent to Router 2. 
    \item The subscribe request from Router 2 is routed to Router 1 by checking the source information at Router 2's RIB. After verifying the request, Router 1's RIB records P2 as the data sink. 
    \item The subscribe request from Router 1 is routed to the robot by checking the source information at Router 1's RIB. If the destination is not found, the router broadcasts a query to other routers. 
    \item The robot's ROS2 publisher sends a ROS2 message to the proxy. The proxy forwards it to Router 1, Router 1 forwards to Router 2, and Router 2 to the cloud subscriber. At each hop, the messages are forwarded from source to sink.
\end{enumerate*}

\subsection{Secure Communication}
\label{sec:design:secure}

\begin{figure}
    \centering
    \includegraphics[width=0.85\linewidth]{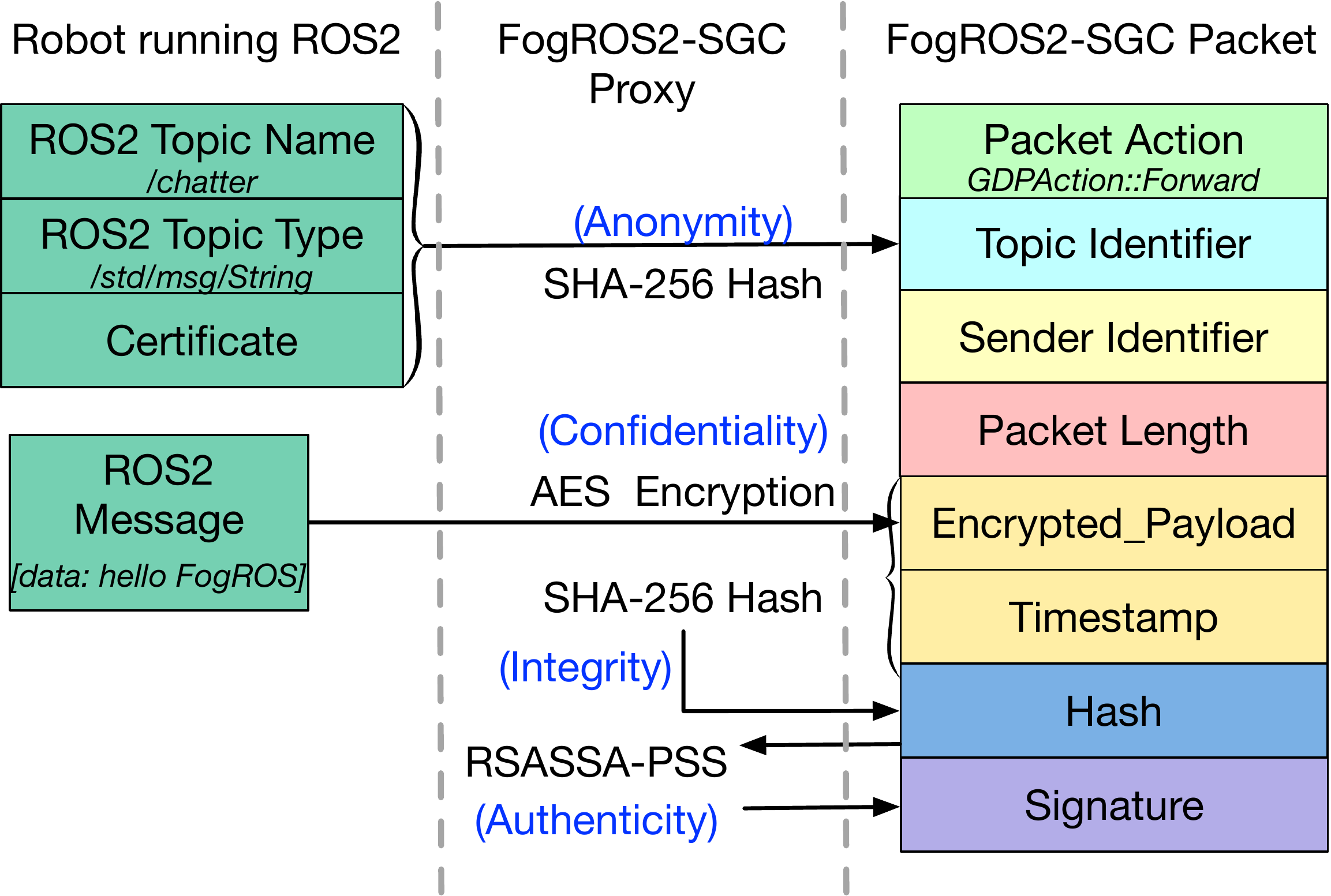}
    \caption{An illustration of the cryptographic tools used by \proxy to protect a ROS2 string message. The \algname Packet on the right is the message that is routed by \algname. The payload is encrypted to protect the confidentiality of the original ROS2 message. The encrypted data is hashed so that the receiver can verify the message is intact. The hash is signed with the sender's key so that the receiver can verify that the message comes from an authentic and authorized sender. 
    }
        \vspace{-15pt}
    \label{fig:design:header}
\end{figure}

The security of the communication is achieved by using a secure network protocol between routers. We use Datagram Transport Layer Security (DTLS)~\cite{dtls_algorithm} to provide communications privacy. The DTLS protocol provides secure and authenticated communication on User Datagram Protocol (UDP) and includes a built-in mechanism for dealing with lost or out-of-order packets. 
DTLS on UDP is well suited for latency-critical robotics communications systems, due to its lightweight nature and low overhead compared to transmission Control Protocl(TCP). The cryptographic algorithms used to secure ROS2 packet generation process can be found in Fig. \ref{fig:design:header}. The message has the following security guarantees: \textbf{Confidentiality:} The ROS2 messages are encrypted with AES Encryption~\cite{daemen1999aes} to ensure that only parties with the correct cryptographic key can decrypt the original ROS2 message data. \textbf{Integrity:} The encrypted message is hashed by SHA-256~\cite{appel2015verification} so the receiver or third-party auditor can easily verify that the message is intact and no other attacker has tampered with the message.  \textbf{Authenticity:} The hashed message is signed by the RSASSA-PSS~\cite{rsassa} algorithm so that receivers can verify that the message is sent from an authorized sender. 

To tailor the security with the communication patterns of robotics applications, \algname allows flexible peering with other routers or end points. One may choose to use a dedicated DTLS connection per ROS2 topic, which is ideal for large message payload and frequent communication (e.g., video streaming). One may also choose to use a shared DTLS tunnel, where multiple ROS2 topics share the same DTLS connection. Sharing the same connection reduces the cost of secure connection management and message processing, which is good for small message payloads and less frequent communication.

\subsection{Transparent and Compatible \proxy} 
\label{sec:design:proxy}
\proxy is the interface between \algname and the ROS2 network. 
In order to allow seamless integration with \textbf{any} unmodified ROS 2 application code and mainstream DDS vendors, 
\proxy converts between ROS2 communication and \algname communication bidirectionally. 
The user first identifies ROS2 topics that they wish to publish or subscribe through a configuration file. The proxy launches a local ROS2 publisher or subscriber for the corresponding topic. New messages from the 
local ROS2 network are actively subscribed to by the proxy, and sent to the \algname network. Once the verified subscribers receive the messages, they convert them to standard ROS messages and publish to their local ROS2 network. 

To allow non-ROS2 programs to communicate with ROS2 nodes, \proxy converts ROS2 messages to a unified JSON-based message format in transit. As a result, \algname can be extended to a variety of protocols such as TCP, UDP, DTLS, TLS, and gRPC. Note, however, that some of the protocols need special handling to be aligned with \algname. For example, gRPC requires the IP addresses of both robot and cloud for bi-directional message passing.

\subsection{Compute and Memory-Efficient \router} 
\label{sec:design:router}
\begin{figure}
    \centering
    \includegraphics[width=0.8\linewidth]{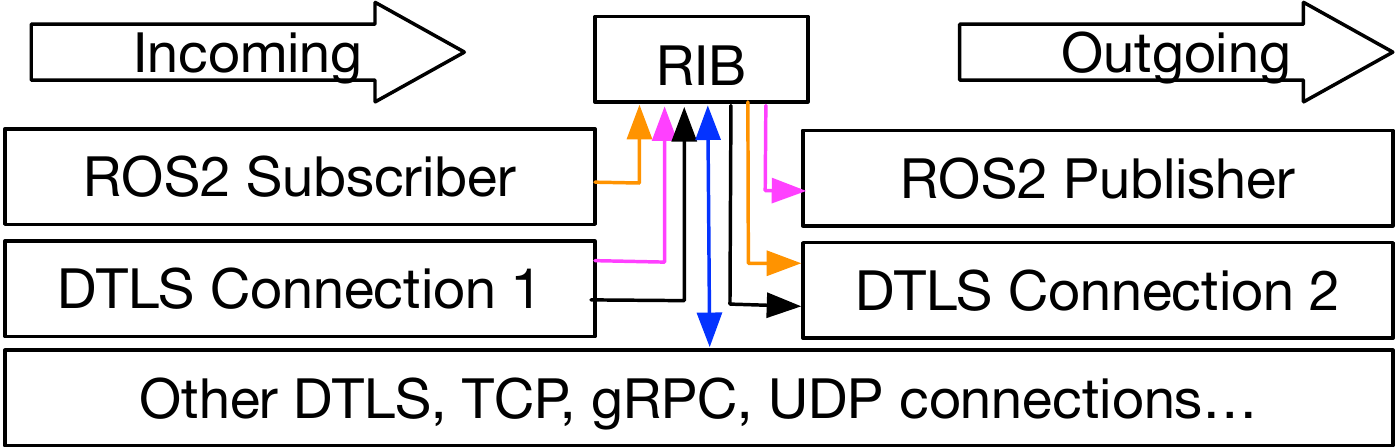}       
    \caption{\router architecture. (\textcolor{orange}{\textbf{Orange}}) Subscribe to a local ROS2 network and publish to \algname routing network. (\textcolor{magenta}{\textbf{Magenta}}) Receive from \algname routing network and publish to local ROS2 network. (\textcolor{black}{\textbf{Black}}) Intermediate \router that facilitates message routing. 
    \router asynchronously reads, writes, and manages all the network connections. All the message passing (arrows) is zero-copy and does not require movement of actual messages. }
    \vspace{-15pt}
    \label{fig:connection-based-arch}
\end{figure}

\algname can be deployed on low-power robots under memory and compute constaints, so an {efficient} implementation of routing algorithm in Section \ref{sec:design:routing} is crucial to the overall performance of the system. Fig. \ref{fig:connection-based-arch} shows an architecture of \router. An idiomatic workflow of the router implementation is to (1) receive data from ROS2/network, (2) decide which network connection to forward, and (3) forward data to ROS2/network. Because \algname needs to be extensible to heterogeneous network protocols, the router needs to maintain many simultaneous network connections, ranging from ROS2's publish/subscribe protocol to general network protocol such as DTLS.

Because low-power robots run under memory constaints, memory copying operations and synchronization operations (such as mutex) are expensive. 
\router is implemented in Rust~\cite{matsakis2014rust} to eliminate memory copying operations and the need for synchronization. Rust is a programming language that features a single ownership model: every data object has a single owner, and passing the data is moving the ownership from one variable to another. As a result, it prevents race conditions and reduces data copying by enforcing the passing of data objects by references instead of values.

Robots with few CPU cores usually have low network performance, because network operations are usually \textit{blocking}, where the entire packet processing halts and waits for the network operations to finish. \algname improves CPU utilization by leveraging asynchronous Rust interfaces~\cite{tokio}. Asynchronous interfaces are non-blocking, removing the network operations out of the critical path of message processing.

\section{Evaluation} 

We evaluate \algname on system benchmarks to show how it performs over alternative designs and on robotics benchmarks to show how robotics applications benefit from \algname. We also demonstrate \algname on four physical robot arms running Fleet-DAgger~\cite{hoque2022fleet}, a multi-robot learning application. 
In Sections~\ref{ssec:sysbench} and \ref{ssec:cloudbench}, we use an Intel NUC with an Intel\textsuperscript\textregistered{} Pentium\textsuperscript\textregistered{} Silver J5005 CPU @ 1.50\,GHz with a 5\,Mbps network connection to act as the robot. The robot is connected with a Standard DS3 v2 cloud instance (4 vCPUs, 14 GiB memory) on Microsoft Azure. The robot is located at California (west coast of US), and the cloud server is located at Virginia (east coast of US).

\subsection{System Benchmarks}\label{ssec:sysbench}

We evaluate the performance of \algname's message processing latency and throughput against other distributed ROS2 systems. %
Messages are sent in binary with type \texttt{sensor\_msgs/CompressedImage} and response with string type \texttt{std\_msgs/String}. 
  We compare against the following baselines 
 (1) \textbf{VPN}: We use Wireguard VPN ~\cite{wireguard}, which is the same VPN as FogROS2 \cite{ichnowski2020fog} 
 (2) \textbf{Rosbridge:} Rosbridge is the most commonly used websocket proxy that allows non-ROS code to interact with ROS code. We use Rosbridge in combination with Rosduct in the same way as in FogROS \cite{chen2021fogros}. 
 (3) \textbf{Capsule:} We use Capsule, a software switch inspired by Netbricks \cite{panda2016netbricks}, to emulate the design of \algname. We also implement rosduct~\cite{rosduct} in ROS2 that converts between ROS2 and network traffic. The detailed description and implementation can be found in FogROS-G~\cite{chen2022fogros}. \textbf{\algname} uses the default DTLS network protocol. We include \textbf{\algname-TCP} that uses TCP instead of DTLS as a variant. 
 
\begin{figure}
    \centering
    \includegraphics[width=0.65\linewidth]{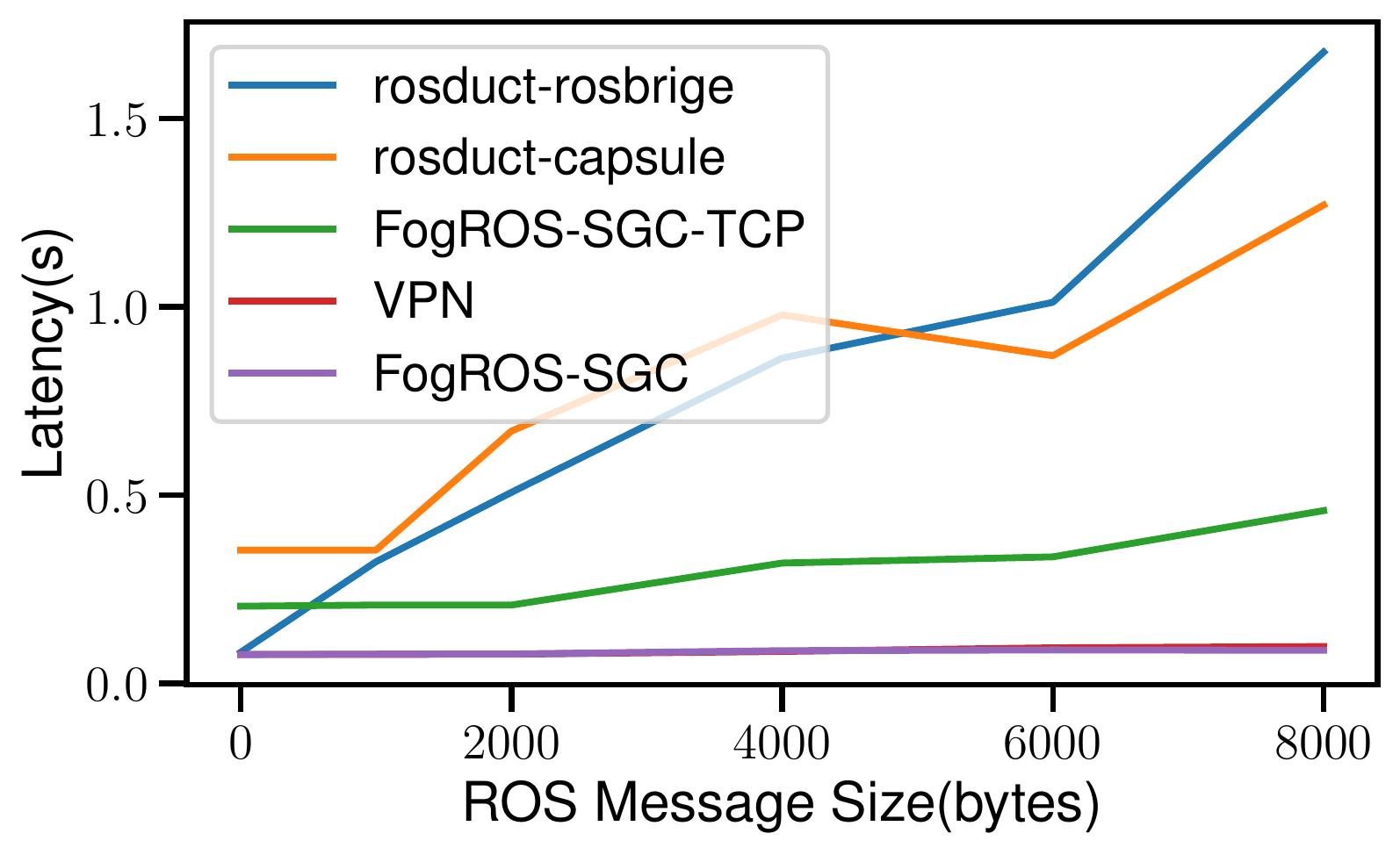}
    \caption{\textbf{Message round trip latency to the cloud} (lower is better). Latency is averaged over more than 50 packet window. \algname is 19 times faster than rosbridge baseline for 8000 byte message. }
    \label{fig:eval:rtt}
    \vspace{-8pt}
\end{figure}

\textbf{ROS2 Message Latency:} We measure the Round Trip Time (RTT) between when a robot publishes a ROS2 message to the cloud and when data is received by the cloud, which echoes a short message on a separate ROS2 topic. The RTT also includes the time of parsing the messages and analyzing the latency. The result can be found in Fig. \ref{fig:eval:rtt}. \algname with DTLS has similar performance as VPN, which has 0.076s round trip latency for small messages. \algname is 10.2\% faster than VPN for 8000 byte messages (0.088 vs 0.097). \algname is 19$\times$ faster than rosduct-rosbridge (0.088 vs 1.67). There are two reasons for this: (1) Rosduct is implemented in Python and provably slower than Rust. It uses blocking network operations while \algname uses non-blocking asynchronous network operations for sending and receiving data. (2) Rosbridge requires seralization of binary messages in JSON, which require more bytes and lead to larger messages. 

\textbf{ROS2 Message Throughput:} Message throughput is measured by the number of messages processed per second. Different from other experiments, throughput is measured on the local area network connected with Ethernet, in order to prevent network bandwidth from being the bottleneck. Table \ref{tab:eval:throughput} shows the message processing throughput. \algname achieves near-native throughput as ROS2 and incurs only 3\% overhead due to the security and conversion to a unified message format. 
\algname has 2.1$\times$ higher throughput than rosbridge, because rosbridge requires more bytes to serialize binary strings.  

\begin{table}
    \centering
    \footnotesize
    \begin{tabular}{ l c } 
\toprule
 Protocol & Throughput (msg/second) \\ 
\midrule
 Original ROS & 330.43 \\
 SROS2 & 320.17 \\ 
 Rosduct-Rosbridge & 152.79 \\ 
 \algname-TCP & 268.03 \\ 
 \algname & 320.40 \\ 
\bottomrule
\end{tabular}
\caption{\textbf{Message throughput evaluation of \algname} (higher is better). Every message is 1000 bytes. The throughput of \algname is near native performance while adding secure and global connectivity and 2.1 times higher than rosbridge.}
\label{tab:eval:throughput}
\vspace{-8pt}
\end{table}

\textbf{Startup and Advertisement Time:} In a RIB that has 10,000 routing records, the average time for publishing a name to the RIB takes 4ms and subscribing to a name from RIB takes 2ms. The average startup time from starting a program to receiving the first message takes 2.4 ms.

\subsection{Cloud Robotics Application Benchmarks}\label{ssec:cloudbench}
We evaluate the network latency of \algname with 3 example cloud robotics applications: SLAM with ORB-SLAM2~\cite{mur2017orb}, Grasp Planning with Dex-Net~\cite{mahler2017dex}, and Motion Planning with Motion Planning Templates (MPT)~\cite{ichnowski2019mpt}.  The detailed description of these benchmarks can be found in~\cite{chen2021fogros}. 

As detailed in Table \ref{tab:latency}, 
although \algname can scale to multiple robots and provide fine grained access control for the robots, it demonstrates even better point-to-point performance than VPN in the vSLAM and grasp planning experiments.
\algname is 9.42 times faster than rosbridge-rosduct on compressed grasp planning images. 
However, \algname cannot reliably transmit large and uncompressed grasp planning matrices. The raw matrix after serialization is larger than 13MB. We observe a significant amount of lost and out of order messages because the default transport protocol of \algname is DTLS over UDP and the communication channel does not recover from lost and out of order messages. Although transmitting such large message within single ROS2 message is rare, users can choose other supported transport protocols (such as TCP, gRPC) to meet the requirement of their applications.

\newcolumntype{R}[2]{%
    >{\adjustbox{angle=#1,lap=\width-(#2)}\bgroup}%
    l%
    <{\egroup}%
}
\newcommand*\rot{\multicolumn{1}{R{45}{1em}}}%

\begin{table}[t]
    \centering
    \footnotesize
    \resizebox{\linewidth}{!}{%
    \begin{tabular}{@{}lrrrrrc@{\quad}r@{}}\toprule
                  & \multicolumn{2}{c}{vSLAM} & \multicolumn{2}{c}{Grasp Planning}  & 
                  \multicolumn{2}{c}{Motion Planning} \\
                  \cmidrule(lr){2-3}
                  \cmidrule(lr){4-5} \cmidrule(lr){6-7}
         Scenario   & \rot{fr1/xyz1} &\rot{fr1/loop} & \rot{raw matrix} & \rot{Compressed}  & \rot{Apartment} &  \rot{Cubicle}   \\
         \midrule
         rosduct-rosbridge  & 10.31 & 10.29 & 20.3 &  13.67 & 0.08 & 0.08   \\
         VPN & 1.16 & 1.45 & \textbf{5.7}  & 1.47 &  0.07  &  0.07 \\
        \algname-TCP &1.19 & 1.57 & 8.4 & 1.58 & 0.07 & 0.07  \\
         \algname & \textbf{1.15} & \textbf{1.42} & - & \textbf{1.45} & 0.07 & 0.07 \\
         \bottomrule
    \end{tabular}
    }
    \caption{\textbf{Network latency of \algname on cloud robotics applications} (lower is better) \algname is better than rosduct-rosbridge and VPN on vSLAM and compressed grasp planning. We conducted motion planning on other scenarios (Home, TwistyCool) and the latency is the same.  }
    \label{tab:latency}
    \vspace{-10pt}
\end{table}

\subsection{Case Study: Fleet-DAgger}
\begin{figure}
    \centering
    \includegraphics[width=\linewidth]{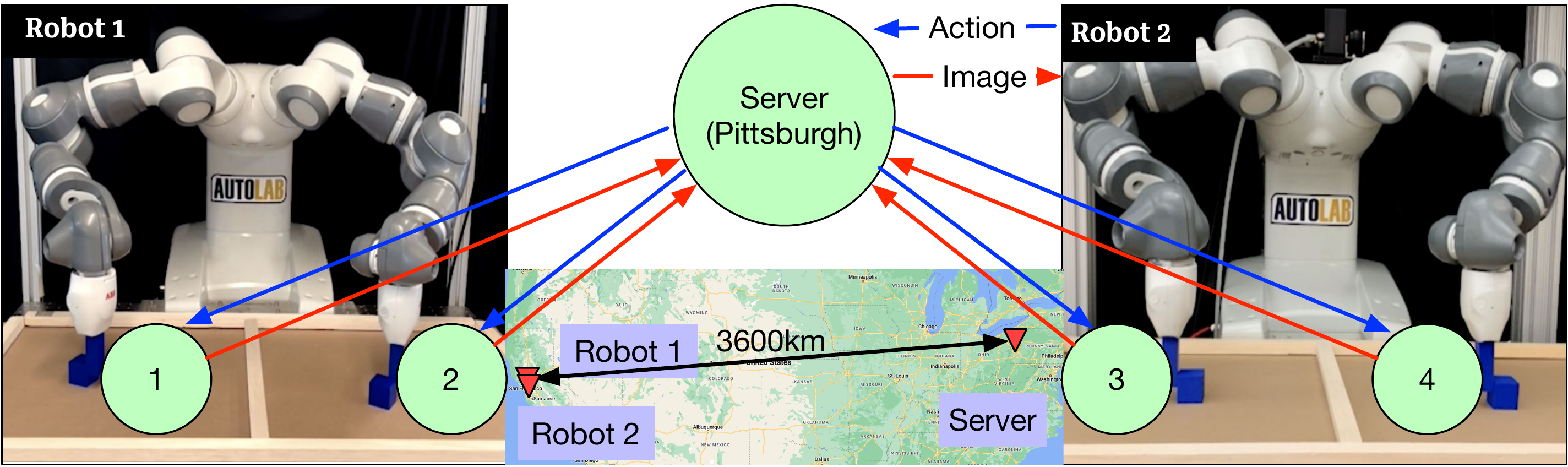}
    \caption{\textbf{The experiment setup of Fleet-DAgger}. Two ABB YuMi robots located in two separate buildings in Berkeley utilize computation from a server located in Pittsburgh for an image based block pushing task.}
    \label{fig:eval:fleet:setup}
    \vspace{-15pt}
\end{figure}
We apply \algname to the control of a fleet of 4 physical robot arms, an increasingly relevant setting in robotics and the third motivating example in Fig. \ref{fig:examples}. We use the physical experiment setup from Fleet-DAgger \cite{hoque2022fleet}, where each robot simultaneously performs an image-based block-pushing task (see Fig.~\ref{fig:examples}C). The task is to repeatedly push a cube to a goal region randomly generated in the image, where a new goal is sampled from the reachable workspace upon reaching the previous goal. The 4 workspaces have an identical setup (but different block positions and goals) to enable the aggregation of each robot's data into a shared dataset and training of a single shared policy on this dataset, as is typical in fleet learning~\cite{hoque2022fleet}. When autonomous control is unreliable, the robots fall back on and learn from remote human teleoperation, where global connectivity can dramatically increase the number of available humans. The arms belong to two bimanual ABB YuMi robots in two different labs about \SI{1}{\km} apart with separate local area networks. To test global connectivity, compute is off-loaded to a separate node in a third local area network at Carnegie Mellon University \SI{3600}{\km} away, where the robot nodes send images of the current state and receive actions to execute.

In a previous implementation, Hoque et al.~\cite{hoque2022fleet} use Secure Shell (SSH) and Secure File Transfer Protocol (SFTP) to communicate between robots and the centralized compute node and Python multiprocessing to enable simultaneous execution. This approach requires storing all SSH credentials at a single node (a security concern), writing image data to the file system of all nodes at every timestep, complex asynchronous programming, and restricting all node locations to within the university campus firewall. To mitigate these issues, we (1) re-implement the communication system with ROS2 and (2) seamlessly connect all nodes with \algname with TCP by modifying only a single configuration text file on each node. Relative to the previous implementation, the \algname implementation reduces communication time by 64\% (Table~\ref{tab:eval:fleetdagger}), where communication time includes image transmission latency and synchronization across all arms but not machine learning or arm motion. \algname also reduces communication time by 33\% relative to the initial implementation even when the robots are in Berkeley, CA and the server is moved to Pittsburgh, PA. Note that the SSH method does not work between Berkeley and Pittsburgh due to university network firewalls~\cite{berkeley_ssh_rule}. %
A diagram of the system architecture is in Fig. \ref{fig:eval:fleet:setup}. 
\begin{table}
    \centering
    \footnotesize
    \begin{tabular}{ c c c } 
\toprule
 Communication System & Server Location & Communication Time (s) \\ 
\midrule
\multirow{2}{*}{SSH + SFTP} & Berkeley, CA & 0.86 \\
& Pittsburgh, PA & - \\
\midrule
\multirow{2}{*}{\algname} & Berkeley, CA & 0.31 \\ 
& Pittsburgh, PA & 0.58\\
\bottomrule
\end{tabular}
\caption{\textbf{Communication time of SSH+SFTP and \algname} (lower is better). \algname with TCP reduces the communication time per experiment step (i.e., one simultaneous action on the 4 arms) by 64\% when compared to SSH+SFTP, and has 33\% lower communication time than SSH+SFTP in Berkeley even if the server is moved to Pittsburgh. SSH does not work if the server is in Pittsburgh due to a university firewall restriction. }
\vspace{-15pt}
\label{tab:eval:fleetdagger}
\end{table}

\section{Conclusions and Future Work}

We present \algname, an extension of FogROS2 that securely connects robotics components across different physical locations and networks.
One limitation of \algname is that users are unable to use retransmission and Quality-of-Service (QoS) mechanisms provided by DDS for inter-ROS2 network communication. However, users can flexibly choose any supported transport protocol (e.g., TCP and gRPC). 
\algname will support other ROS2 communication patterns (e.g., services and actions) in the future. %

\section*{Acknowledgement}
This research was performed at the AUTOLAB at UC Berkeley in affiliation with the Berkeley AI Research (BAIR) Lab. The authors were supported in part by donations from  Bosch, Toyota Research Institute, Google, Siemens, and Autodesk and by equipment grants from PhotoNeo, NVidia, and Intuitive Surgical. The research is also supported by C3.ai Digital Transformation Institute for AI resilience.

\bibliographystyle{IEEEtran}
\bibliography{IEEEabrv,references}

\end{document}